\documentclass[letter, twocolumn]{article}
\usepackage{arxiv}  
\usepackage{times}  
\usepackage{helvet}  
\usepackage{courier} 
\usepackage[hyphens]{url} 
\usepackage{graphicx} 
\usepackage{natbib}  
\usepackage{caption}
\usepackage{algorithm}
\usepackage{algorithmic}
\usepackage{newfloat}
\usepackage{listings}
\DeclareCaptionStyle{ruled}{labelfont=normalfont,labelsep=colon,strut=off}
\floatstyle{ruled}
\newfloat{listing}{tb}{lst}{}
\floatname{listing}{Listing}

\usepackage{amsmath}
\usepackage{amssymb}  
\usepackage{nicefrac, xfrac}

\usepackage{xspace}
\newcommand{\drmd}{\textsc{DRMD}\xspace}

\usepackage{booktabs}
\usepackage{xcolor}
\definecolor{darkgreen}{RGB}{0, 100, 0}
\definecolor{darkred}{RGB}{139, 0, 0}
\definecolor{darkyellow}{RGB}{204, 153, 0}
\usepackage{multirow}
\usepackage{hhline}
\newcommand{\etal}{{\it et al.}}

\title{\textsc{DRMD}: Deep Reinforcement Learning for Malware Detection under Concept Drift}

\author{
    {\bfseries Shae McFadden}\textsuperscript{\rm  $\dagger$ \rm $\ddagger$ \rm  $\mathsection$}, \
    {\bfseries Myles Foley}\textsuperscript{\rm $\ddagger$}, \
    {\bfseries Mario D'Onghia}\textsuperscript{\rm  $\mathsection$}, \\ [0.5ex]
    {\bfseries Chris Hicks}\textsuperscript{\rm $\ddagger$}, \
    {\bfseries Vasilios Mavroudis}\textsuperscript{\rm $\ddagger$}, \
    {\bfseries Nicola Paoletti}\textsuperscript{\rm  $\dagger$}, \ 
    {\bfseries Fabio Pierazzi}\textsuperscript{\rm  $\mathsection$} \\ [0.5ex]
    \textsuperscript{\rm $\dagger$}King's College London, \
    \textsuperscript{\rm $\ddagger$}The Alan Turing Institute, \
    \textsuperscript{\rm $\mathsection$}University College London \\
}

\begin{document}

\maketitle

\begin{abstract}
Malware detection in real-world settings must deal with evolving threats, limited labeling budgets, and uncertain predictions. Traditional classifiers, without additional mechanisms, struggle to maintain performance under concept drift in malware domains, as their supervised learning formulation cannot optimize when to defer decisions to manual labeling and adaptation. Modern malware detection pipelines combine classifiers with monthly active learning (AL) and rejection mechanisms to mitigate the impact of concept drift. In this work, we develop a novel formulation of malware detection as a one-step Markov Decision Process and train a deep reinforcement learning (DRL) agent, simultaneously optimizing sample classification performance and rejecting high-risk samples for manual labeling. We evaluated the joint detection and drift mitigation policy learned by the DRL-based Malware Detection (\drmd) agent through time-aware evaluations on Android malware datasets subject to realistic drift requiring multi-year performance stability. The policies learned under these conditions achieve a higher Area Under Time (AUT) performance compared to standard classification approaches used in the domain, showing improved resilience to concept drift. 
Specifically, the \drmd agent achieved an average AUT improvement of $8.66$ and $10.90$ for the classification-only and classification-rejection policies, respectively.
Our results demonstrate for the first time that DRL can facilitate effective malware detection and improved resiliency to concept drift in the dynamic setting of Android malware detection.

\textbf{AAAI-26}: This is an extended version of the paper \textit{DRMD: Deep Reinforcement Learning for Malware Detection under Concept Drift} published at AAAI 2026.

\end{abstract}

\textbf{Code}: \url{https://github.com/s2labres/DRMD}

\section{Introduction}

    Malware poses a significant and ever‐evolving threat to mobile devices, personal computers, and enterprise systems. Every day, tens of thousands of new Android and Windows applications appear, far outpacing the capacity for manual analysis. 
    Automated malware detection using machine learning approaches is therefore crucial to ensure user security. 
    However, unlike domains that can directly process raw data, malware detection relies on vectorized abstractions to represent applications for classification. Therefore, feature extraction plays a pivotal role in determining the overall performance of a malware detection system. As a result, malware representations have been addressed using various feature spaces~\citep{Drebin,MaMaDroid,zhang2020enhancing,ramda}, each incorporating different information (e.g. API calls or app permissions) into feature vectors for malware detection classifiers. 

    Traditional supervised classifiers struggle to maintain malware detection performance in isolation as benign and malicious behaviors evolve over time~\citep{Tesseract,miller2016reviewer,allix2015your}. While classification techniques from domains such as computer vision are effective on stationary data, they struggle to adapt to the dynamic nature of malware that actively seeks to evade detection~\citep{kan2024tesseract}. Approaches such as active learning (AL) to periodically retrain on a subset of new samples~\citep{settles2009active, chen2023continuous} and classification with rejection to selectively abstain from classification~\citep{Transcend,Transcendent,CADE} can mitigate performance degradation caused by evolving threats.  

    In this work, we unify classification, AL, and rejection within a single decision framework using deep reinforcement learning (DRL). Prior work from Coscia \etal~\citeyear{coscia2024sinner} extended the Imbalanced Classification MDP (ICMDP)~\cite{lin2020deep} for Windows PE malware family classification to address class imbalance. While achieving impressive performance, prior work~\cite{coscia2024sinner,binxiang2019deep} does not unify classification and rejection in a single decision process, nor does it consider concept drift introduced by malware evolution. Furthermore, episodes in the ICMDP formulation involve classifying multiple samples: this introduces invalid sequential dependencies through the state transitions between samples. 
     
    We present the first approach in this domain that reformulates malware detection as a \emph{one-step MDP}, or \emph{contextual bandit}, where each episode corresponds to a single sample, thus avoiding spurious dependencies between states. Leveraging this formulation, we train a DRL-based Malware Detection (\drmd) agent that directly optimizes the long-term trade-off between detection accuracy, labeling cost, and misclassification risk under realistic concept drift. The policies learned by \drmd provide statistically significant gains over the best baseline in 75\% of settings evaluated across different: feature spaces, datasets, and drift scenarios. These results demonstrate that \drmd is a promising enhancement to current malware detection pipelines. In summary, this paper makes the following key contributions:
    \begin{itemize}
      \item \textit{Malware Detection MDP.} We introduce a novel one‐step MDP formulation of malware detection that treats each sample independently, named the MD-MDP. This rectifies the spurious dependencies between samples present in prior work.

      \item \textit{Integrated Rejection.} We extend the MD-MDP with a rejection action and reward structure, which enables learning a policy that balances classification accuracy and misclassification risk.
      
      \item \textit{Time-Aware Evaluations.} We demonstrate that the policies learned from the MD-MDP consistently outperform Android malware detection approaches considered across multiple feature spaces and datasets in time-aware evaluations, achieving a relative AUT gain of $8.66$ and $10.90$ for the classification-only and classification-rejection policies, respectively.
    \end{itemize}

\begin{figure*}
    \centering
    \includegraphics[width=1\textwidth]{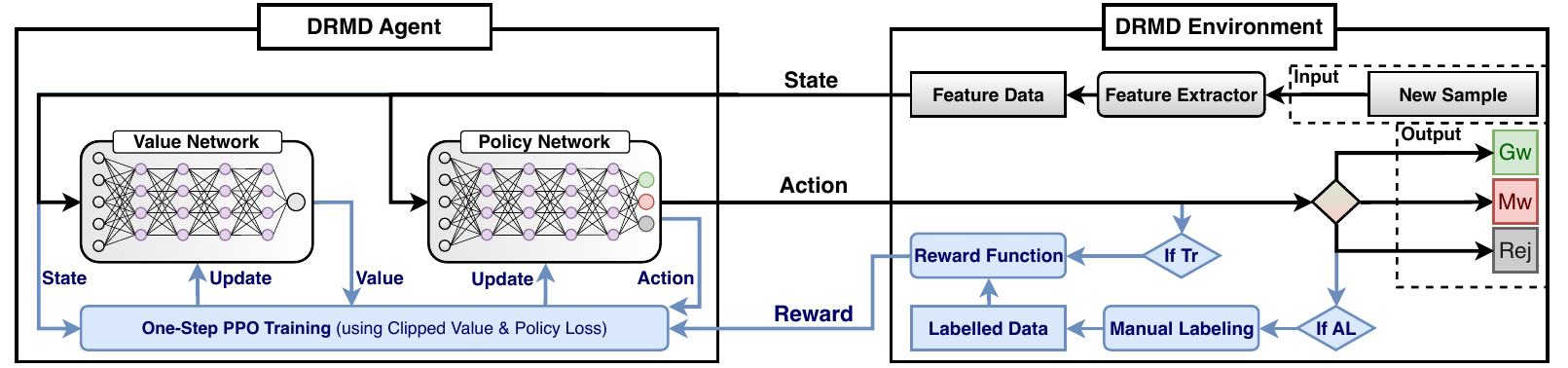}
    \caption{Overview figure showing the one-step classification of a sample using \drmd, with training components in blue.}
    \label{fig:DRMD}
\end{figure*}

\section{Related Work} \label{sec:related_work}
    DRL has demonstrated excellent performance in a variety of decision-making and classification tasks. Lin \etal~\citeyear{lin2020deep} formulated ICMDP, forming states as samples, actions as predicted labels, rewarding +1 for correct and -1 for incorrect classification of minority class samples with majority class sample rewards down-scaled according to the class distribution.
    The use of a Deep Q-Network (DQN) for ICMDP (DQNimb) has been successful on images in the healthcare domain, including stroke detection~\citep{zuo2024stroke}, COVID-19 screening~\citep{yang2024deep}, and other healthcare tasks~\citep{zhou2021deep,jayaprakash2023meddqn,usha2023drl}. However, the application of DRL to malware detection is underexplored. SINNER~\cite{coscia2024sinner} builds on the ICMDP framework and extends its reward function to the multi-class setting, employing a Dueling Double DQN to classify the family of Windows PE malware samples. Earlier work from Binxiang \etal~\citeyear{binxiang2019deep} similarly applies DQN to detect Windows PE malware. Although these approaches improve performance (c.f. Lin \etal~\citeyear{lin2020deep}) by harnessing DRL, they do not consider concept drift, a core problem in malware detection. Furthermore, these approaches are not evaluated in conjunction with AL or rejection mechanisms. In contrast, our work combines classification, rejection, and AL for an agent to learn a single policy to effectively detect malware over time.

\section{Android Malware Detection} \label{sec:background}
A core challenge of malware detection arises from the scale of new samples that emerge daily and the limited capacity for manual review. Onwuzurike \etal~\citeyear{MaMaDroid} estimated that an upper bound of 10,000 applications are submitted to the Google Play Store each day, while Miller \etal~\citeyear{miller2016reviewer} assumed the manual review capacity of an average company at only 80 samples per day. This significant gap necessitates the use of machine learning (ML) to accurately and consistently classify new applications based on previously labeled data.

    \paragraph{Concept Drift} 
        A key limitation of traditional Supervised Learning (SL) classifiers is their reliance on the assumption that data follows an Independent and Identically Distributed (IID) structure~\citep{bishop2006pattern}. However, this assumption does not hold in the domain of Android malware, due to the continuous evolution of both benign applications (goodware) and malware over time~\citep{Tesseract}. Goodware evolves naturally as software developers introduce new features, while malware adapts to better mimic goodware and evade detection. This continual evolution leads to a divergence between training and testing data that expands over time, a phenomenon known as concept drift~\citep{moreno2012unifying}. This performance degradation in malware classifiers poses a significant threat to the security of end users~\citep{allix2015your, miller2016reviewer, Tesseract, kan2024tesseract}.

    \paragraph{AL and Rejection Mechanisms}
        One of the primary strategies to mitigate concept drift is \textit{active learning} (AL), which involves selecting a subset of new samples for manual labeling and periodic retraining~\citep{settles2009active}. The strategy for selecting the most informative samples is the key differentiator between selection approaches. However, manual labeling in the malware domain is resource-intensive. Therefore, AL techniques must carefully balance selecting the most informative samples while minimizing labeling costs. In contrast, rejection strategies identifies samples with a high probability of misclassification. These rejected samples are quarantined for manual inspection or other investigate processes. Both AL and rejection mechanisms in Android malware detection are typically employed periodically to mitigate concept drift, with common evaluations considering monthly periods for these mechanisms~\cite{kan2024tesseract, chen2023continuous, mcfadden2024impact}. In this work, we focus on the use of uncertainty-based AL and rejection~\citep{settles2009active}. The integration of other methods, such as conformal evaluation~\cite{Transcend,Transcendent} or explanation-based approaches~\cite{CADE}, into the reward design of DRL-based detectors presents an interesting but non-trival direction for future work. This paper aims to establish a foundational basis for DRL-based malware detection integrating AL and rejection.

    \paragraph{Time-Aware Evaluation} \label{sec:timeaware}
        Since concept drift in malware detection is inherently time-dependent, time-aware evaluations are essential to assess classifier performance in real-world scenarios. A valid time-aware evaluation must adhere to the following constraints~\citep{Tesseract,kan2024tesseract}. \textbf{(C1)} Training samples must always precede testing samples. \textbf{(C2)} All testing samples must be drawn from the same time window. \textbf{(C3)} The malware distribution in the testing set must reflect a realistic distribution. Violating these constraints can introduce biases that artificially inflate the performance of the classifier, leading to unrealistic expectations in practical deployments. In order to evaluate the performance of detectors, we use the Area Under Time (AUT) metric, introduced by Pendlebury \etal\citeyear{Tesseract}, and defined as:
        \begin{equation}
        \displaystyle AUT(P,N) = \frac{1}{N-1} \sum_{k=1}^{N-1}  \frac{[P(X_{k+1}) + P(X_k)]}{2} \label{eq:aut}
        \end{equation}
        where: $P(X_k)$ is the performance on samples $X_k$ using metric $P$, with $P$ being the $F_1$ score in our paper.

\section{\drmd Design} \label{sec:mdp_formulation}
    \drmd is a DRL-based approach that, in its most complete form, unifies classification, AL, and rejection in a single malware detection pipeline. \drmd first formulates the Malware Detection MDP (MD-MDP), then uses a deep neural network to learn an effective detection policy. Concretely, MD-MDP extends ICMDP~\cite{lin2020deep} with corrected episode definitions and integrates the mechanisms necessary for effective malware detection in practice. See Figure \ref{fig:DRMD} for an overview of \drmd.

    \paragraph{Motivation}
    The application of DRL to detection tasks has been shown to improve performance in more complex domains with class imbalance~\cite{zhou2021deep,zuo2024stroke,yang2024deep}. However, the use of DRL in malware detection has seen little work~\cite{coscia2024sinner,binxiang2019deep} likely due to the drift and class imbalance of the domain. A policy for malware detection should consider the larger decision-making process of classification, rejection, and AL; thus allowing for the optimization of performance to be rejection-aware. Finally, the agent must adapt to concept drift within the domain constraints. As a result, DRL policies must move beyond the stationary perspectives of prior problem domains. Our MD-MDP and agent design achieve these considerations through a one-step episodic formulation that leverages an expanded reward function for temporal robustness and the integration of rejection.

    \subsection{MD-MDP}\label{sec:mdmdp}
    Prior work has applied the ICMDP to several domains such as malware family classification~\cite{coscia2024sinner}. However, the formulation of ICMDP assumes state transitions between classification samples, thus treating independent samples as sequentially dependent. To overcome this, we cast malware detection as a one‐step MDP (\textit{aka} contextual bandit). 
    
    \paragraph{Episodes}\label{sec:episodes}
    In our formulation, each episode consists of a single step: classifying or rejecting one sample. This corrects the state transitions introduced by the ICMDP formulation by isolating each sample into its own episode. Therefore, our MD-MDP formulation ensures that policy updates do not introduce spurious correlations as a result of artificial dependencies between independent samples.

    \paragraph{States}
    For episode $e$, the sample $x^e$ consists of $d$ features, where $F \subseteq \mathbb{R}^d$ represents the feature space, hence $x^e \in F$. The state consists of the features of the current sample; therefore, the state space and state are defined as $S=F$ and $s = x^e$, respectively.

    \paragraph{Actions}
    A chosen action $a \in A$ represents the decision of the agent for the current state $s$. The action space is $A = \{a_{gw},a_{mw},a_{rej}\}$, where $a_{gw}$ classifies the sample as goodware, $a_{mw}$ as malware and $a_{rej}$ rejects the sample, abstaining from classification. \textit{Active learning} is integrated by retraining on samples for which the agent selected $a_{rej}$.

    \paragraph{Rewards}
        The rewards need to encapsulate the dynamics of classifications and rejection. The three underlying components of classification in the domain are the accuracy of the predictions, the natural imbalance in the distribution of the samples, and the evolution of samples over time. 
 
        For \textit{prediction accuracy}, correct classifications receive a positive reward and incorrect classifications incur a penalty. Given the true label $y^e \in \{0,1\}$ for a state $y(s^e)=y^e$, the accuracy reward is: 
        \begin{equation}
            R_{\mathrm{acc}}(s,a) =
                \begin{cases}
                +1 & \text{if } a = y(s), \\
                -1 & \text{if } a \neq y(s).
                \end{cases}
        \end{equation}
        
        For \textit{class imbalance}, since Android malware samples are the minority class ($\hat\sigma\approx10\%$)~\citep{Tesseract}, the rewards and penalties for malware samples ($y(s)=1$) are proportionally increased. We use upscaling instead of majority class reward downscaling used by Lin \etal~\citeyear{lin2020deep} as we introduce additional scaling factors for temporal and rejection values. Therefore, the malware reward upscaling is defined as:
        \begin{equation}
            R_{imb}(s) = \begin{cases}
                \nicefrac{1}{\hat\sigma} & \text{if } y(s) = 1, \\
                1 & \text{if } y(s) = 0.
                \end{cases}
        \end{equation}

        For \textit{temporal robustness}, we upscale the rewards for the current state $s$ by it relative temporal position $T_{pos}$, in months, compared to the the first training state $s^0$, such that: 
        
        \begin{equation}
            R_{tmp}(s) = \nicefrac{1}{2}\max\left(1,\ T_{pos}(s)-T_{pos}(s^0)\right)
        \end{equation}
        
        For \textit{rejection}, the reward is comprised of a flat rejection cost ($R_{cost}$), the inverted reward of the next most likely action ($R_{nml}=R_{clf}(s,a\neq a_{rej})$), as well as a rejection scaling factor based on the value of $R_{nml}$ and $\hat\sigma$. The scaling factor balances the rewards for classifying and rejecting malware predictions while incentivizing rejection of uncertain goodware predictions. Together, these components penalize interference with correct classifications and incentivize rejection of incorrect classifications. The rejection reward is defined as:
        \begin{equation} 
            R_{rej}(s, a) = 
                \begin{cases} 
                    - R_{cost} - \frac{1}{\hat\sigma}R_{nml}  & \text{if }R_{nml} \le 0 \\
                    - R_{cost} - \hat\sigma R_{nml} & \text{else. }
                \end{cases} 
        \end{equation}
        
        The classification ($R_{clf}$) and complete \drmd ($R_{cr}$) reward functions are defined as:
        \begin{equation}
            R_{clf}(s, a) = R_{acc}(s,a) \times R_{tmp}(s) \times R_{imb}(s)
        \end{equation}
        \begin{equation} 
            R_{cr}(s, a) = 
                \begin{cases} 
                    R_{clf}(s, a)  & \text{if } a \in \{a_{gw}, a_{mw}\} \\
                    R_{rej}(s, a) & \text{if } a = a_{rej}
                \end{cases} 
        \end{equation}

    Using our formulation, we create two types of policies for evaluation in Section~\ref{sec:results}. First, a Classification-Only Policy ($\pi_{clf}$) using the classification actions ($a_{gw}$ and $a_{mw}$), and reward $R_{clf}(s, a)$. Second, a Classification-Rejection Policy ($\pi_{cr}$) that uses all actions, and reward $R_{cr}(s, a)$.
    
    \subsection{Agent Architecture} \label{sec:arch}
    We use Proximal Policy Optimization (PPO)~\citep{schulman2017proximal}, modified for one-step episodes, to train the \drmd agent. As the clipped update and actor-critic architecture of PPO allows efficient scaling to high-dimensional environments while maintaining training stability~\citep{nguyen2021deep}. Additionally, PPO has seen successful application to several other security domains such as automated cyber defense~\cite{foleyCAGEI22, foleyCAGEII22, hicksCAGE3_23, bates2023reward}, vulnerability discovery~\cite{mcfadden2024wendigo, foley2025apirl}, and adversarial machine learning~\cite{tsingenopoulos2022captcha,tsingenopoulos2024train}. The on-policy nature of PPO and its use of rollout updates facilitate learning on the most recent samples compared to experience replay buffers, used in off-policy methods, which would be more susceptible to concept drift. 
    
    While we have formulated the MD-MDP as a one-step MDP or contextual bandit problem, existing deep contextual bandits (DCBs) are not well-suited to our setting. First, DCBs assume stationary sub-Gaussian (bounded) reward noise~\cite{zhou2020neural}, an assumption violated in malware domains due to concept drift. Existing work has explored the use of DCBs to facilitate AL~\cite{tae_falcon_2024,ban_neural_2024} but not the application of AL to DCBs.
    
    The agent employs separate actor and critic neural networks, each consisting of four hidden layers with $512$ neurons per layer, using LeakyReLU activations and dropout for regularization. The full architecture and hyperparameters are detailed in the Appendix \ref{app:hyper}. During training, each epoch consists of one episode per training sample using the current policy. The collected states, actions, and rewards are then used to compute the clipped policy and value losses to update the actor and critic networks over mini batches. When new samples become available after the initial training, the agent will be fine-tuned on a sliding window of the newest samples (instead of on all samples) to focus updates on adapting to changes in the domain.

\section{Experimental Settings} \label{sec:experimental_settings}

    \paragraph{Datasets}
    We use two Android malware datasets in our evaluation: Transcendent~\citep{Transcendent} and Hypercube~\citep{chow2025breaking}. 
    Transcendent provides an established Android malware dataset~\cite{kan2024tesseract,mcfadden2024impact,herzog2025aurora}, while Hypercube is an updated dataset with newer malware samples. Transcendent consists of $259,230$ applications collected between $2014$ and $2018$. Hypercube contains $159,839$ applications between $2021$ and $2023$. For both datasets, we use the first year for training and subsequent years for testing, allowing evaluation of performance under concept drift. Both datasets contain a malware distribution of $\approx10\%$ according to the spatial constraint of Tesseract~\citep{Tesseract}, and the samples have been labeled using a standard VirusTotal detection threshold ($VTT = 4$), according to the recommendations of \cite{Tesseract,kan2024tesseract}.  

    \begin{table*}[t]
            \centering
            \caption{\textbf{Classification-Only Policy.} The AUT performance of \drmd across different settings, datasets, and feature spaces are reported alongside the performance differences compared to the best baseline ($\Delta$Base), the ICMDP ($\Delta$IC), and the DCBs ($\Delta$DCB). All results are conducted over five runs and paired t-tests are used for statistical significance testing (* $p < 0.05$, ** $p < 0.01$, *** $p < 0.001$).
            }
            \label{tab:post-hoc-results}
            \resizebox{1\textwidth}{!}{
            \begin{tabular}{ll|llll|llll|llll|llll}
                \toprule
                \multirow{2}{*}{AL} & \multirow{2}{*}{Rej}  & \multicolumn{4}{c|}{Hypercube-Drebin}  & \multicolumn{4}{c|}{Hypercube-Ramda}  & \multicolumn{4}{c|}{Transcendent-Drebin}  & \multicolumn{4}{c}{Transcendent-Ramda} \\
                 &  & AUT & $\Delta$Base & $\Delta$IC & $\Delta$DCB & AUT & $\Delta$Base & $\Delta$IC & $\Delta$DCB & AUT & $\Delta$Base & $\Delta$IC & $\Delta$DCB & AUT & $\Delta$Base & $\Delta$IC & $\Delta$DCB \\
                \midrule
                0 & 0 & 63.33$\,\pm\,$1.14 & \textcolor{darkred}{-1.15} & \textcolor{darkgreen}{+0.88} & \textcolor{darkgreen}{+5.15}$^{**}$ & 45.03$\,\pm\,$0.68 & \textcolor{darkgreen}{+13.74}$^{***}$ & \textcolor{gray}{+0.36} & \textcolor{darkgreen}{+4.19}$^{***}$ & 58.07$\,\pm\,$0.53 & \textcolor{darkgreen}{+8.46}$^{***}$ & \textcolor{darkgreen}{+1.11} & \textcolor{darkgreen}{+10.52}$^{***}$ & 50.46$\,\pm\,$0.87 & \textcolor{darkgreen}{+6.72}$^{**}$ & \textcolor{gray}{+0.10} & \textcolor{darkgreen}{+13.05}$^{***}$ \\
                0 & 50 & 65.05$\,\pm\,$0.98 & \textcolor{darkgreen}{+0.66} & \textcolor{gray}{+0.48} & \textcolor{darkgreen}{+5.88}$^{**}$ & 45.84$\,\pm\,$0.86 & \textcolor{darkgreen}{+18.69}$^{***}$ & \textcolor{gray}{+0.31} & \textcolor{darkgreen}{+4.12}$^{***}$ & 59.44$\,\pm\,$0.54 & \textcolor{darkgreen}{+11.48}$^{***}$ & \textcolor{gray}{+0.49} & \textcolor{darkgreen}{+9.98}$^{***}$ & 52.22$\,\pm\,$1.00 & \textcolor{darkgreen}{+9.64}$^{**}$ & \textcolor{gray}{+0.20} & \textcolor{darkgreen}{+13.07}$^{***}$ \\
                0 & 100 & 67.13$\,\pm\,$1.07 & \textcolor{darkgreen}{+3.59}$^{**}$ & \textcolor{gray}{+0.34} & \textcolor{darkgreen}{+6.94}$^{**}$ & 46.87$\,\pm\,$0.77 & \textcolor{darkgreen}{+26.65}$^{***}$ & \textcolor{gray}{+0.20} & \textcolor{darkgreen}{+4.33}$^{***}$ & 59.91$\,\pm\,$0.58 & \textcolor{darkgreen}{+11.88}$^{***}$ & \textcolor{darkgreen}{+0.50} & \textcolor{darkgreen}{+10.43}$^{***}$ & 52.55$\,\pm\,$0.91 & \textcolor{darkgreen}{+10.41}$^{**}$ & \textcolor{gray}{+0.37} & \textcolor{darkgreen}{+13.54}$^{***}$ \\
                0 & 200 & 70.48$\,\pm\,$0.91 & \textcolor{darkgreen}{+6.42}$^{***}$ & \textcolor{gray}{+0.14} & \textcolor{darkgreen}{+8.47}$^{**}$ & 48.89$\,\pm\,$0.79 & \textcolor{darkgreen}{+29.77}$^{***}$ & \textcolor{darkgreen}{+0.63} & \textcolor{darkgreen}{+4.94}$^{***}$ & 61.92$\,\pm\,$0.62 & \textcolor{darkgreen}{+14.53}$^{***}$ & \textcolor{darkgreen}{+0.57} & \textcolor{darkgreen}{+10.93}$^{***}$ & 53.41$\,\pm\,$1.07 & \textcolor{darkgreen}{+13.66}$^{***}$ & \textcolor{gray}{-0.12} & \textcolor{darkgreen}{+12.44}$^{***}$ \\
                0 & 400 & 75.02$\,\pm\,$0.80 & \textcolor{darkgreen}{+9.98}$^{***}$ & \textcolor{gray}{-0.43} & \textcolor{darkgreen}{+10.22}$^{***}$ & 52.50$\,\pm\,$0.66 & \textcolor{darkgreen}{+32.92}$^{***}$ & \textcolor{darkgreen}{+1.60} & \textcolor{darkgreen}{+6.02}$^{***}$ & 63.26$\,\pm\,$0.87 & \textcolor{darkgreen}{+18.29}$^{***}$ & \textcolor{gray}{+0.29} & \textcolor{darkgreen}{+10.61}$^{***}$ & 55.39$\,\pm\,$0.86 & \textcolor{darkgreen}{+21.66}$^{***}$ & \textcolor{darkgreen}{+1.02} & \textcolor{darkgreen}{+13.43}$^{***}$ \\
                \midrule
                50 & 0 & 74.05$\,\pm\,$1.35 & \textcolor{darkgreen}{+6.01}$^{**}$ & \textcolor{darkgreen}{+5.56}$^{**}$ & \textcolor{darkgreen}{+15.13}$^{***}$ & 46.20$\,\pm\,$0.79 & \textcolor{darkgreen}{+8.55}$^{***}$ & \textcolor{gray}{+0.45} & \textcolor{darkgreen}{+4.30}$^{***}$ & 70.65$\,\pm\,$0.92 & \textcolor{gray}{+0.04} & \textcolor{darkgreen}{+5.31}$^{***}$ & \textcolor{darkgreen}{+12.61}$^{***}$ & 57.74$\,\pm\,$2.42 & \textcolor{darkgreen}{+7.28}$^{**}$ & \textcolor{gray}{+0.30} & \textcolor{darkgreen}{+11.30}$^{***}$ \\
                50 & 50 & 76.06$\,\pm\,$1.59 & \textcolor{darkgreen}{+7.91}$^{***}$ & \textcolor{darkgreen}{+5.63}$^{**}$ & \textcolor{darkgreen}{+15.67}$^{***}$ & 46.37$\,\pm\,$0.74 & \textcolor{darkgreen}{+11.69}$^{***}$ & \textcolor{gray}{-0.33} & \textcolor{darkgreen}{+3.47}$^{***}$ & 73.00$\,\pm\,$0.87 & \textcolor{darkgreen}{+3.96}$^{*}$ & \textcolor{darkgreen}{+5.10}$^{***}$ & \textcolor{darkgreen}{+13.29}$^{***}$ & 58.29$\,\pm\,$2.51 & \textcolor{darkgreen}{+8.76}$^{**}$ & \textcolor{gray}{+0.04} & \textcolor{darkgreen}{+10.05}$^{***}$ \\
                50 & 100 & 77.36$\,\pm\,$1.66 & \textcolor{darkgreen}{+9.23}$^{***}$ & \textcolor{darkgreen}{+5.13}$^{**}$ & \textcolor{darkgreen}{+15.90}$^{***}$ & 46.86$\,\pm\,$0.87 & \textcolor{darkgreen}{+16.00}$^{***}$ & \textcolor{gray}{-0.19} & \textcolor{darkgreen}{+3.03}$^{***}$ & 73.04$\,\pm\,$0.84 & \textcolor{darkgreen}{+2.26} & \textcolor{darkgreen}{+4.74}$^{***}$ & \textcolor{darkgreen}{+12.91}$^{***}$ & 59.89$\,\pm\,$2.42 & \textcolor{darkgreen}{+10.31}$^{**}$ & \textcolor{gray}{+0.38} & \textcolor{darkgreen}{+11.36}$^{***}$ \\
                50 & 200 & 79.16$\,\pm\,$1.77 & \textcolor{darkgreen}{+10.78}$^{**}$ & \textcolor{darkgreen}{+4.12}$^{**}$ & \textcolor{darkgreen}{+15.86}$^{***}$ & 48.14$\,\pm\,$0.75 & \textcolor{darkgreen}{+21.29}$^{***}$ & \textcolor{darkgreen}{+0.55} & \textcolor{darkgreen}{+2.86}$^{***}$ & 73.53$\,\pm\,$1.92 & \textcolor{darkgreen}{+3.58}$^{*}$ & \textcolor{darkgreen}{+3.22}$^{*}$ & \textcolor{darkgreen}{+12.07}$^{***}$ & 60.20$\,\pm\,$2.00 & \textcolor{darkgreen}{+12.63}$^{***}$ & \textcolor{gray}{-0.27} & \textcolor{darkgreen}{+10.05}$^{***}$ \\
                50 & 400 & 81.89$\,\pm\,$1.85 & \textcolor{darkgreen}{+13.13}$^{***}$ & \textcolor{darkgreen}{+2.84}$^{*}$ & \textcolor{darkgreen}{+15.85}$^{***}$ & 51.23$\,\pm\,$1.13 & \textcolor{darkgreen}{+23.97}$^{***}$ & \textcolor{darkgreen}{+2.35}$^{**}$ & \textcolor{darkgreen}{+2.94}$^{***}$ & 74.82$\,\pm\,$0.97 & \textcolor{darkgreen}{+4.43}$^{**}$ & \textcolor{darkgreen}{+3.10}$^{*}$ & \textcolor{darkgreen}{+12.04}$^{***}$ & 62.90$\,\pm\,$2.08 & \textcolor{darkgreen}{+18.61}$^{***}$ & \textcolor{gray}{+0.30} & \textcolor{darkgreen}{+11.46}$^{***}$ \\
                \midrule
                100 & 0 & 75.94$\,\pm\,$0.93 & \textcolor{darkgreen}{+5.56}$^{***}$ & \textcolor{darkgreen}{+3.15} & \textcolor{darkgreen}{+15.10}$^{***}$ & 48.27$\,\pm\,$0.67 & \textcolor{darkgreen}{+7.73}$^{***}$ & \textcolor{darkgreen}{+1.45}$^{*}$ & \textcolor{darkgreen}{+5.72}$^{***}$ & 71.06$\,\pm\,$0.98 & \textcolor{darkred}{-0.74}$^{**}$ & \textcolor{darkgreen}{+4.50}$^{**}$ & \textcolor{darkgreen}{+10.71}$^{***}$ & 59.77$\,\pm\,$1.60 & \textcolor{darkgreen}{+4.08}$^{**}$ & \textcolor{darkgreen}{+1.66} & \textcolor{darkgreen}{+10.88}$^{***}$ \\
                100 & 50 & 77.96$\,\pm\,$0.99 & \textcolor{darkgreen}{+6.74}$^{***}$ & \textcolor{darkgreen}{+2.95} & \textcolor{darkgreen}{+15.76}$^{***}$ & 49.09$\,\pm\,$0.71 & \textcolor{darkgreen}{+11.18}$^{***}$ & \textcolor{darkgreen}{+1.15}$^{*}$ & \textcolor{darkgreen}{+5.63}$^{***}$ & 73.32$\,\pm\,$1.13 & \textcolor{darkgreen}{+2.40}$^{*}$ & \textcolor{darkgreen}{+4.81}$^{**}$ & \textcolor{darkgreen}{+11.26}$^{***}$ & 60.60$\,\pm\,$1.88 & \textcolor{darkgreen}{+6.89}$^{**}$ & \textcolor{darkgreen}{+1.87} & \textcolor{darkgreen}{+10.08}$^{***}$ \\
                100 & 100 & 79.28$\,\pm\,$1.15 & \textcolor{darkgreen}{+8.23}$^{***}$ & \textcolor{darkgreen}{+2.55} & \textcolor{darkgreen}{+15.94}$^{***}$ & 49.73$\,\pm\,$0.77 & \textcolor{darkgreen}{+15.63}$^{***}$ & \textcolor{darkgreen}{+1.16}$^{*}$ & \textcolor{darkgreen}{+5.32}$^{***}$ & 73.51$\,\pm\,$1.01 & \textcolor{darkgreen}{+1.45} & \textcolor{darkgreen}{+4.30}$^{*}$ & \textcolor{darkgreen}{+10.96}$^{***}$ & 61.92$\,\pm\,$1.58 & \textcolor{darkgreen}{+8.01}$^{**}$ & \textcolor{darkgreen}{+2.02} & \textcolor{darkgreen}{+11.10}$^{***}$ \\
                100 & 200 & 81.10$\,\pm\,$1.47 & \textcolor{darkgreen}{+10.53}$^{***}$ & \textcolor{darkgreen}{+2.01} & \textcolor{darkgreen}{+16.06}$^{***}$ & 51.04$\,\pm\,$0.80 & \textcolor{darkgreen}{+23.68}$^{***}$ & \textcolor{darkgreen}{+1.74}$^{**}$ & \textcolor{darkgreen}{+4.96}$^{***}$ & 74.17$\,\pm\,$1.52 & \textcolor{darkgreen}{+3.31}$^{*}$ & \textcolor{darkgreen}{+2.86} & \textcolor{darkgreen}{+10.69}$^{***}$ & 62.22$\,\pm\,$1.45 & \textcolor{darkgreen}{+10.49}$^{***}$ & \textcolor{darkgreen}{+0.74} & \textcolor{darkgreen}{+9.75}$^{***}$ \\
                100 & 400 & 83.61$\,\pm\,$1.68 & \textcolor{darkgreen}{+12.27}$^{***}$ & \textcolor{darkgreen}{+1.00} & \textcolor{darkgreen}{+15.78}$^{***}$ & 53.97$\,\pm\,$1.13 & \textcolor{darkgreen}{+25.74}$^{***}$ & \textcolor{darkgreen}{+2.98}$^{**}$ & \textcolor{darkgreen}{+5.01}$^{***}$ & 74.90$\,\pm\,$1.13 & \textcolor{darkgreen}{+3.80}$^{**}$ & \textcolor{darkgreen}{+1.98} & \textcolor{darkgreen}{+9.51}$^{***}$ & 65.03$\,\pm\,$1.46 & \textcolor{darkgreen}{+16.82}$^{***}$ & \textcolor{darkgreen}{+1.76} & \textcolor{darkgreen}{+11.09}$^{***}$ \\
                \midrule
                200 & 0 & 76.51$\,\pm\,$0.41 & \textcolor{darkgreen}{+2.42}$^{*}$ & \textcolor{darkgreen}{+3.22}$^{**}$ & \textcolor{darkgreen}{+13.89}$^{***}$ & 49.15$\,\pm\,$0.95 & \textcolor{darkgreen}{+6.08}$^{***}$ & \textcolor{darkgreen}{+1.22} & \textcolor{darkgreen}{+6.03}$^{***}$ & 72.67$\,\pm\,$0.38 & \textcolor{darkred}{-2.73}$^{***}$ & \textcolor{darkgreen}{+4.45}$^{***}$ & \textcolor{darkgreen}{+10.39}$^{***}$ & 60.14$\,\pm\,$0.32 & \textcolor{gray}{+0.06} & \textcolor{darkgreen}{+0.75} & \textcolor{darkgreen}{+8.91}$^{***}$ \\
                200 & 50 & 78.56$\,\pm\,$0.25 & \textcolor{darkgreen}{+4.76}$^{***}$ & \textcolor{darkgreen}{+3.16}$^{**}$ & \textcolor{darkgreen}{+14.24}$^{***}$ & 50.12$\,\pm\,$0.88 & \textcolor{darkgreen}{+8.43}$^{***}$ & \textcolor{darkgreen}{+1.05} & \textcolor{darkgreen}{+6.06}$^{***}$ & 74.98$\,\pm\,$0.54 & \textcolor{gray}{+0.02} & \textcolor{darkgreen}{+4.53}$^{***}$ & \textcolor{darkgreen}{+11.22}$^{***}$ & 60.78$\,\pm\,$0.48 & \textcolor{darkgreen}{+2.12}$^{*}$ & \textcolor{darkgreen}{+0.96} & \textcolor{darkgreen}{+7.72}$^{***}$ \\
                200 & 100 & 79.74$\,\pm\,$0.30 & \textcolor{darkgreen}{+6.15}$^{***}$ & \textcolor{darkgreen}{+2.88}$^{**}$ & \textcolor{darkgreen}{+14.26}$^{***}$ & 50.96$\,\pm\,$0.76 & \textcolor{darkgreen}{+12.44}$^{***}$ & \textcolor{darkgreen}{+1.24} & \textcolor{darkgreen}{+5.94}$^{***}$ & 75.22$\,\pm\,$0.51 & \textcolor{darkred}{-1.06} & \textcolor{darkgreen}{+4.24}$^{**}$ & \textcolor{darkgreen}{+10.84}$^{***}$ & 62.64$\,\pm\,$0.23 & \textcolor{darkgreen}{+3.87}$^{***}$ & \textcolor{darkgreen}{+1.29}$^{*}$ & \textcolor{darkgreen}{+9.27}$^{***}$ \\
                200 & 200 & 81.43$\,\pm\,$0.42 & \textcolor{darkgreen}{+8.07}$^{***}$ & \textcolor{darkgreen}{+2.57}$^{**}$ & \textcolor{darkgreen}{+14.04}$^{***}$ & 52.26$\,\pm\,$0.78 & \textcolor{darkgreen}{+20.58}$^{***}$ & \textcolor{darkgreen}{+1.56}$^{*}$ & \textcolor{darkgreen}{+5.66}$^{***}$ & 76.38$\,\pm\,$0.52 & \textcolor{darkgreen}{+1.85} & \textcolor{darkgreen}{+3.65}$^{**}$ & \textcolor{darkgreen}{+10.63}$^{***}$ & 62.58$\,\pm\,$0.26 & \textcolor{darkgreen}{+6.18}$^{***}$ & \textcolor{gray}{+0.11} & \textcolor{darkgreen}{+7.51}$^{***}$ \\
                200 & 400 & 83.86$\,\pm\,$0.36 & \textcolor{darkgreen}{+10.21}$^{***}$ & \textcolor{darkgreen}{+2.11}$^{**}$ & \textcolor{darkgreen}{+13.69}$^{***}$ & 55.10$\,\pm\,$0.81 & \textcolor{darkgreen}{+27.48}$^{***}$ & \textcolor{darkgreen}{+2.35}$^{**}$ & \textcolor{darkgreen}{+5.54}$^{***}$ & 76.65$\,\pm\,$0.66 & \textcolor{darkgreen}{+1.08} & \textcolor{darkgreen}{+2.94}$^{**}$ & \textcolor{darkgreen}{+9.59}$^{***}$ & 65.50$\,\pm\,$0.25 & \textcolor{darkgreen}{+14.52}$^{***}$ & \textcolor{darkgreen}{+1.00}$^{**}$ & \textcolor{darkgreen}{+8.80}$^{***}$ \\
                \midrule
                400 & 0 & 77.45$\,\pm\,$0.38 & \textcolor{darkgreen}{+1.42} & \textcolor{darkgreen}{+3.26}$^{***}$ & \textcolor{darkgreen}{+12.77}$^{***}$ & 51.01$\,\pm\,$0.54 & \textcolor{darkgreen}{+5.50}$^{***}$ & \textcolor{darkgreen}{+2.53}$^{**}$ & \textcolor{darkgreen}{+7.08}$^{***}$ & 72.50$\,\pm\,$1.53 & \textcolor{darkred}{-8.33}$^{***}$ & \textcolor{darkgreen}{+3.41}$^{*}$ & \textcolor{darkgreen}{+8.87}$^{***}$ & 60.81$\,\pm\,$0.76 & \textcolor{darkred}{-2.59}$^{**}$ & \textcolor{darkgreen}{+0.79} & \textcolor{darkgreen}{+7.75}$^{***}$ \\
                400 & 50 & 79.66$\,\pm\,$0.32 & \textcolor{darkgreen}{+2.56}$^{*}$ & \textcolor{darkgreen}{+3.20}$^{***}$ & \textcolor{darkgreen}{+13.46}$^{***}$ & 52.27$\,\pm\,$0.67 & \textcolor{darkgreen}{+8.61}$^{***}$ & \textcolor{darkgreen}{+2.53}$^{**}$ & \textcolor{darkgreen}{+7.34}$^{***}$ & 74.79$\,\pm\,$1.58 & \textcolor{darkred}{-5.16}$^{**}$ & \textcolor{darkgreen}{+3.36}$^{*}$ & \textcolor{darkgreen}{+9.56}$^{***}$ & 62.12$\,\pm\,$1.34 & \textcolor{gray}{+0.16} & \textcolor{darkgreen}{+1.57} & \textcolor{darkgreen}{+7.10}$^{***}$ \\
                400 & 100 & 80.90$\,\pm\,$0.38 & \textcolor{darkgreen}{+3.46}$^{**}$ & \textcolor{darkgreen}{+2.87}$^{**}$ & \textcolor{darkgreen}{+13.43}$^{***}$ & 53.19$\,\pm\,$0.71 & \textcolor{darkgreen}{+12.55}$^{***}$ & \textcolor{darkgreen}{+2.63}$^{**}$ & \textcolor{darkgreen}{+7.29}$^{***}$ & 75.07$\,\pm\,$1.68 & \textcolor{darkred}{-6.30}$^{*}$ & \textcolor{darkgreen}{+2.81} & \textcolor{darkgreen}{+9.17}$^{***}$ & 63.34$\,\pm\,$0.95 & \textcolor{darkgreen}{+1.19} & \textcolor{darkgreen}{+1.06} & \textcolor{darkgreen}{+7.91}$^{***}$ \\
                400 & 200 & 82.53$\,\pm\,$0.48 & \textcolor{darkgreen}{+5.21}$^{**}$ & \textcolor{darkgreen}{+2.59}$^{**}$ & \textcolor{darkgreen}{+13.13}$^{***}$ & 54.58$\,\pm\,$0.72 & \textcolor{darkgreen}{+20.29}$^{***}$ & \textcolor{darkgreen}{+2.85}$^{**}$ & \textcolor{darkgreen}{+7.02}$^{***}$ & 76.04$\,\pm\,$1.75 & \textcolor{darkred}{-4.04} & \textcolor{darkgreen}{+2.16} & \textcolor{darkgreen}{+8.62}$^{***}$ & 63.46$\,\pm\,$0.74 & \textcolor{darkgreen}{+2.98}$^{**}$ & \textcolor{gray}{+0.09} & \textcolor{darkgreen}{+6.23}$^{***}$ \\
                400 & 400 & 84.66$\,\pm\,$0.52 & \textcolor{darkgreen}{+7.67}$^{**}$ & \textcolor{darkgreen}{+1.98}$^{*}$ & \textcolor{darkgreen}{+12.54}$^{***}$ & 56.94$\,\pm\,$0.47 & \textcolor{darkgreen}{+29.07}$^{***}$ & \textcolor{darkgreen}{+3.11}$^{**}$ & \textcolor{darkgreen}{+6.15}$^{***}$ & 76.43$\,\pm\,$1.72 & \textcolor{darkred}{-3.69} & \textcolor{darkgreen}{+1.65} & \textcolor{darkgreen}{+7.74}$^{***}$ & 66.39$\,\pm\,$0.80 & \textcolor{darkgreen}{+10.68}$^{***}$ & \textcolor{darkgreen}{+1.15} & \textcolor{darkgreen}{+7.60}$^{***}$ \\
                \bottomrule
            \end{tabular}
            }
        \end{table*}
    
    \paragraph{Feature Spaces}
    To ensure that the performance of DRMD is independent of a single feature representation, we evaluate it using two different Android malware feature spaces \emph{for each dataset}. We use feature spaces that enable us to evaluate our approach across varying amounts of information per sample. First, we use the \textit{Drebin}~\cite{Drebin} feature space, consisting of a sparse, high-dimensional binary vector, capturing static features extracted from the application's metadata and bytecode, including permissions, API calls, and network addresses. For our experiments, we follow \citeauthor{kan2024tesseract} in selecting the $10,000$ most informative Drebin features as it maintains performance while managing computational complexity. This representation has become the standard in Android malware detection~\citep{Tesseract,chen2023continuous,chow2025breaking}. The second feature space that we use is \textit{Ramda}~\cite{ramda}, a compact feature space designed for adversarial robustness, which focuses on three critical static attributes: permissions, intent actions, and sensitive API calls. Each application is encoded as a 379-dimensional binary vector that captures the presence or absence of these features.

    \paragraph{Baselines}
    To compare our novel MD-MDP with the ICMDP, we evaluate the two model designs 
    using the same (our) DRL agent architecture, thereby isolating the effect of this factor on the performance. The difference in performance is denoted as $\Delta IC$. Moreover, we compare DRMD with a version of our model architecture that uses Supervised Learning (SL) and standard cross-entropy loss, which we call SL-DRMD. In addition to SL-DRMD, for baselines, we also evaluate commonly used classifiers in Android malware detection: Drebin, DeepDrebin, and Ramda. The Drebin~\citep{Drebin} classifier is a linear support vector machine (SVM) originally used for its corresponding feature space and serves as a foundational benchmark due to its simplicity. DeepDrebin~\citep{deepDrebin} applies a multilayer perceptron (MLP), with two hidden layer of 200 neurons, and represents a classical deep learning approach to Android malware detection without extensive modification. Finally, the Ramda~\citep{ramda} classifier, developed for its corresponding feature space, uses a variational autoencoder (VAE) and a MLP to classify samples. Together, Drebin, DeepDrebin, Ramda, and SL-DRMD are evaluated across five random seeds on the four dataset-feature-space pairs and the difference between \drmd and the best performing baseline, for that setting, is reported as $\Delta Base$. Beyond $\Delta IC$ and $\Delta Base$, we also compare our agent architecture against two DCBs, NeuralTS~\cite{zhang2020neural} and NeuralUCB~\cite{zhou2020neural}. $\Delta DCB$ reports the performance difference between \drmd and the best performing DCB using our state, action, and rewards. The hyperparameters for all of the baselines are provided in Appendix~\ref{app:hyper}.

    \section{Evaluation} \label{sec:results}
    This evaluation aims not only to investigate the base performance of \drmd but also to assess the impact that including rejection and AL (both monthly and integrated) has on model performance over time. We evaluate \drmd using five seeds, following the work of Alam \etal~\citeyear{alam2024revisiting}, for each combination of dataset, feature space, and configuration, totaling 172 settings. Thus, we mitigate the impact of stochastic variance by capturing performance trends over multiple runs and across a wide range of conditions. Our results demonstrate that \drmd, both classification-only and classification-rejection policies, consistently outperform considered baselines, achieving an average AUT improvement of $8.66$ and $10.90$, respectively. Furthermore, \drmd significantly outperforms DCBs, NeuralTS and NeuralUCB, over all settings for the classification–only policy with an average $\Delta Base$ of $9.77$.

    \subsection{Classification–Only Policy}\label{sec:class_only_pol}
        First, we consider DRMD trained to learn a Classification–Only policy ($\pi_{clf}$), without the rejection action. Initially, when there is no AL or rejection (the first row of Table~\ref{tab:post-hoc-results}), \drmd surpasses the best baselines by $6.94$ on average across five runs and all dataset-feature-space pairs. These initial results are promising, yet do not consider a complete malware detection pipeline. Thus, we consider malware detection pipeline configurations using monthly budgets for the uncertainty-based AL and rejection (Rej). 

        \begin{table*}[t]
            \centering
            \caption{\textbf{Classification-Rejection Policy.} \drmd performance and average monthly rejection/selection rates across cost levels. 
            $\Delta Base$ uses the seed-wise average rejections as the monthly rejection budget. All results are conducted as in Table \ref{tab:post-hoc-results}.
            }
            \label{tab:integrated-rejection}
            \resizebox{1\textwidth}{!}{
            \begin{tabular}{c|c|llr|llr|llr|llr}
                \toprule
                \multirow{5}{*}{$IR$} & \multirow{2}{*}{$R_{cost}$} & \multicolumn{3}{c|}{Hypercube-Drebin} & \multicolumn{3}{c|}{Hypercube-Ramda} & \multicolumn{3}{c|}{Transcendent-Drebin} & \multicolumn{3}{c}{Transcendent-Ramda} \\
                & & \multicolumn{1}{l}{AUT $\pm$ std} & \multicolumn{1}{l}{$\Delta$Base} & \multicolumn{1}{r|}{Rejected} & \multicolumn{1}{l}{AUT $\pm$ std} & \multicolumn{1}{l}{$\Delta$Base} & \multicolumn{1}{r|}{Rejected} & \multicolumn{1}{l}{AUT $\pm$ std} & \multicolumn{1}{l}{$\Delta$Base} & \multicolumn{1}{r|}{Rejected} & \multicolumn{1}{l}{AUT $\pm$ std}  & \multicolumn{1}{l}{$\Delta$Base} & \multicolumn{1}{r}{Rejected} \\
                \cline{2-14}
                & 0  & 61.68$\,\pm\,$1.71 & \textcolor{darkred}{-2.46} & 52.02  & 50.54$\,\pm\,$2.82 & \textcolor{darkgreen}{+30.69}$^{***}$ & 533.80  & 60.70$\,\pm\,$2.34 & \textcolor{darkgreen}{+14.08}$^{**}$ & 217.31  & 56.90$\,\pm\,$0.74 & \textcolor{darkgreen}{+23.98}$^{***}$ & 599.47 \\
                & -0.1  & 63.17$\,\pm\,$1.27 & \textcolor{darkred}{-0.90} & 58.17  & 50.99$\,\pm\,$4.94 & \textcolor{darkgreen}{+31.14}$^{***}$ & 529.27  & 58.51$\,\pm\,$4.34 & \textcolor{darkgreen}{+11.67}$^{*}$ & 203.50  & 55.84$\,\pm\,$1.09 & \textcolor{darkgreen}{+22.91}$^{***}$ & 595.97 \\
                & -1  & 63.19$\,\pm\,$2.51 & \textcolor{darkred}{-1.18} & 38.98  & 48.18$\,\pm\,$3.91 & \textcolor{darkgreen}{+28.83}$^{***}$ & 353.58  & 58.96$\,\pm\,$1.97 & \textcolor{darkgreen}{+11.39}$^{***}$ & 130.99  & 56.45$\,\pm\,$1.36 & \textcolor{darkgreen}{+23.67}$^{***}$ & 541.41 \\
                \midrule \midrule
                \multirow{4}{*}{$IRAL$} & $R_{cost}$ & \multicolumn{1}{l}{AUT $\pm$ std} & \multicolumn{1}{l}{$\Delta$Base} & \multicolumn{1}{r|}{Rej\&AL} & \multicolumn{1}{l}{AUT $\pm$ std} & \multicolumn{1}{l}{$\Delta$Base} & \multicolumn{1}{r|}{Rej\&AL} & \multicolumn{1}{l}{AUT $\pm$ std} & \multicolumn{1}{l}{$\Delta$Base} & \multicolumn{1}{r|}{Rej\&AL} & \multicolumn{1}{l}{AUT $\pm$ std} & \multicolumn{1}{l}{$\Delta$Base} & \multicolumn{1}{r}{Rej\&AL} \\
                \cline{2-14}
                & 0  & 69.75$\,\pm\,$3.13 & \textcolor{darkgreen}{+4.48}$^{**}$ & 28.95  & 64.35$\,\pm\,$2.26 & \textcolor{darkgreen}{+33.84}$^{***}$ & 956.81  & 66.71$\,\pm\,$3.20 & \textcolor{darkred}{-0.91} & 40.26  & 68.10$\,\pm\,$1.73 & \textcolor{darkgreen}{+12.00}$^{***}$ & 506.85 \\
                & -0.1  & 67.51$\,\pm\,$5.41 & \textcolor{darkgreen}{+3.82} & 21.99  & 65.39$\,\pm\,$1.82 & \textcolor{darkgreen}{+36.65}$^{***}$ & 956.98  & 60.39$\,\pm\,$8.24 & \textcolor{darkred}{-2.10} & 20.98  & 68.32$\,\pm\,$6.43 & \textcolor{darkgreen}{+14.65}$^{*}$ & 869.61 \\
                & -1  & 64.61$\,\pm\,$4.42 & \textcolor{darkgreen}{+1.00} & 24.09  & 57.41$\,\pm\,$4.47 & \textcolor{darkgreen}{+29.98}$^{***}$ & 489.02  & 57.73$\,\pm\,$5.70 & \textcolor{darkred}{-2.18} & 9.71  & 63.62$\,\pm\,$0.96 & \textcolor{darkgreen}{+7.96}$^{***}$ & 211.16 \\
                \midrule \midrule 
                \multirow{13}{*}{$IRAAL$} & $R_{cost}$ & \multicolumn{1}{l}{AUT $\pm$ std} & \multicolumn{1}{l}{$\Delta$Base} & \multicolumn{1}{r|}{AL$|$Rej} & \multicolumn{1}{l}{AUT $\pm$ std} & \multicolumn{1}{l}{$\Delta$Base} & \multicolumn{1}{r|}{AL$|$Rej} & \multicolumn{1}{l}{AUT $\pm$ std} & \multicolumn{1}{l}{$\Delta$Base} & \multicolumn{1}{r|}{AL$|$Rej} & \multicolumn{1}{l}{AUT $\pm$ std} & \multicolumn{1}{l}{$\Delta$Base} & \multicolumn{1}{r}{AL$|$Rej} \\
                \cline{2-14}
                & \multirow{4}{*}{0}                  & 72.37$\,\pm\,$2.69 & \textcolor{darkgreen}{+5.00}$^{**}$ & 50$|$23.12                & 44.50$\,\pm\,$2.25 & \textcolor{darkgreen}{+17.97}$^{***}$ & 50$|$360.80                & 70.93$\,\pm\,$1.91 & \textcolor{darkgreen}{+2.41} & 50$|$50.07                & 61.09$\,\pm\,$2.01 & \textcolor{darkgreen}{+13.51}$^{***}$ & 50$|$242.63 \\
                &                 & 72.05$\,\pm\,$4.99 & \textcolor{darkgreen}{+3.34} & 100$|$15.34                & 50.86$\,\pm\,$3.17 & \textcolor{darkgreen}{+22.98}$^{***}$ & 100$|$448.57                & 72.69$\,\pm\,$0.29 & \textcolor{darkgreen}{+0.86} & 100$|$38.87                & 63.94$\,\pm\,$0.90 & \textcolor{darkgreen}{+13.68}$^{***}$ & 100$|$279.51 \\
                &                 & 75.98$\,\pm\,$1.95 & \textcolor{darkgreen}{+3.23}$^{***}$ & 200$|$13.37                & 56.23$\,\pm\,$0.84 & \textcolor{darkgreen}{+27.36}$^{***}$ & 200$|$587.93                & 73.11$\,\pm\,$1.22 & \textcolor{darkred}{-1.49}$^{*}$ & 200$|$57.27                & 64.95$\,\pm\,$0.44 & \textcolor{darkgreen}{+13.53}$^{***}$ & 200$|$383.50 \\
                &                 & 78.13$\,\pm\,$1.21 & \textcolor{darkgreen}{+1.95}$^{*}$ & 400$|$18.32                & 60.10$\,\pm\,$1.49 & \textcolor{darkgreen}{+31.21}$^{***}$ & 400$|$724.25                & 74.66$\,\pm\,$0.53 & \textcolor{darkred}{-6.53}$^{**}$ & 400$|$70.01                & 68.10$\,\pm\,$3.31 & \textcolor{darkgreen}{+13.39}$^{**}$ & 400$|$466.65 \\
                \cline{2-14}
                & \multirow{4}{*}{-0.1}                  & 69.68$\,\pm\,$5.69 & \textcolor{darkgreen}{+4.73}$^{*}$ & 50$|$13.67                & 43.84$\,\pm\,$2.45 & \textcolor{darkgreen}{+17.38}$^{***}$ & 50$|$344.86                & 69.74$\,\pm\,$1.36 & \textcolor{darkgreen}{+1.07} & 50$|$34.48                & 61.15$\,\pm\,$1.90 & \textcolor{darkgreen}{+12.51}$^{***}$ & 50$|$227.24 \\
                &                 & 75.29$\,\pm\,$1.78 & \textcolor{darkgreen}{+4.03}$^{**}$ & 100$|$12.72                & 50.87$\,\pm\,$1.88 & \textcolor{darkgreen}{+23.24}$^{***}$ & 100$|$404.13                & 72.49$\,\pm\,$0.68 & \textcolor{darkgreen}{+0.99} & 100$|$39.14                & 63.23$\,\pm\,$1.69 & \textcolor{darkgreen}{+12.96}$^{***}$ & 100$|$286.79 \\
                &                 & 76.80$\,\pm\,$0.64 & \textcolor{darkgreen}{+3.66}$^{***}$ & 200$|$12.91                & 56.09$\,\pm\,$2.14 & \textcolor{darkgreen}{+27.48}$^{***}$ & 200$|$553.63                & 73.76$\,\pm\,$0.75 & \textcolor{darkred}{-0.55} & 200$|$60.88                & 65.10$\,\pm\,$1.55 & \textcolor{darkgreen}{+12.76}$^{***}$ & 200$|$338.65 \\
                &                 & 77.48$\,\pm\,$1.95 & \textcolor{darkgreen}{+1.95}$^{*}$ & 400$|$13.83                & 59.61$\,\pm\,$0.64 & \textcolor{darkgreen}{+30.96}$^{***}$ & 400$|$684.16                & 74.57$\,\pm\,$0.56 & \textcolor{darkred}{-6.63}$^{***}$ & 400$|$64.27                & 66.82$\,\pm\,$1.49 & \textcolor{darkgreen}{+10.65}$^{**}$ & 400$|$401.47 \\
                \cline{2-14}
                & \multirow{4}{*}{-1}                  & 66.84$\,\pm\,$4.29 & \textcolor{darkgreen}{+3.74}$^{*}$ & 50$|$4.18                & 44.99$\,\pm\,$2.34 & \textcolor{darkgreen}{+19.27}$^{***}$ & 50$|$192.70                & 65.04$\,\pm\,$3.35 & \textcolor{darkgreen}{+1.56} & 50$|$6.58                & 60.17$\,\pm\,$1.52 & \textcolor{darkgreen}{+10.65}$^{***}$ & 50$|$82.58 \\
                &                 & 69.18$\,\pm\,$4.36 & \textcolor{darkgreen}{+3.74}$^{*}$ & 100$|$3.71                & 48.26$\,\pm\,$1.44 & \textcolor{darkgreen}{+20.91}$^{***}$ & 100$|$211.69                & 69.03$\,\pm\,$3.58 & \textcolor{darkred}{-1.32} & 100$|$12.30                & 63.56$\,\pm\,$0.99 & \textcolor{darkgreen}{+9.70}$^{***}$ & 100$|$140.76 \\
                &                 & 70.00$\,\pm\,$5.12 & \textcolor{darkgreen}{+2.35} & 200$|$3.73                & 53.45$\,\pm\,$1.20 & \textcolor{darkgreen}{+26.06}$^{***}$ & 200$|$322.32                & 70.54$\,\pm\,$1.43 & \textcolor{darkred}{-2.20} & 200$|$9.04                & 65.30$\,\pm\,$0.90 & \textcolor{darkgreen}{+9.97}$^{***}$ & 200$|$235.29 \\
                &                 & 68.82$\,\pm\,$6.23 & \textcolor{darkred}{-2.04} & 400$|$4.86                & 57.53$\,\pm\,$1.21 & \textcolor{darkgreen}{+29.31}$^{***}$ & 400$|$503.93                & 72.56$\,\pm\,$1.12 & \textcolor{darkred}{-4.57}$^{***}$ & 400$|$13.76                & 65.15$\,\pm\,$1.17 & \textcolor{darkgreen}{+4.96}$^{**}$ & 400$|$239.47 \\
                \bottomrule
            \end{tabular}
            }
        \end{table*}

        \paragraph{Monthly Rejection}
            Here, we consider how integrating uncertainty-based monthly rejection affects performance when no AL is present. \drmd shows statistically significant improvements in performance over the best baseline in 15 of the 16 settings (rows 2-5 of Table~\ref{tab:post-hoc-results}), with an average $\Delta Base$ of $15.01$. Increasing the number of uncertain samples rejected leads to increased performance for both AUT and $\Delta Base$. This means that the samples \drmd has high certainty on lead to more correct classifications than the baselines. Showing how the use of RL improves both overall accuracy and prediction certainty in malware detection.

        \paragraph{Monthly Active Learning}
            In the AL setting, both the \drmd policy and the baselines use monthly uncertainty-based AL to select samples for retraining at each testing month. \drmd demonstrates statistically significant improvements in 9 of the 16 settings (where $AL > 0$ and $Rej = 0$ in Table~\ref{tab:post-hoc-results}), achieving an average $\Delta Base$ of $2.52$. As in the rejection-only setting, both \drmd and the baselines use the same AL mechanism and budgets. Note that the diminishing returns of AL as performance improves~\cite{mcfadden2024impact, mcfadden2023poster} coupled with the initial performance gains of \drmd contribute to the decreasing $\Delta Base$ as AL rates increase. 
    
        \paragraph{Monthly Rejection \& Active Learning}
            To represent a complete malware detection pipeline we combine both monthly AL and rejection in this evaluation, shown in Table~\ref{tab:post-hoc-results} where $AL > 0$ and $Rej > 0$. \drmd has statistically significant improvements in $52$ of the $64$ settings, achieving up to $29.07$ AUT improvement ($AL=400$ and $Rej=400$), and an average $\Delta Base$ of $8.71$. Thus, when the rejection of samples at high risk of misclassification and the cost-sensitive drift adaptation of AL are applied together on \drmd, this performs better than when only using either individually, highlighting how our approach can yield improvements in real world settings.

        \paragraph{MD-MDP vs ICMDP}
            To isolate the difference from MD-MDP compared to ICMDP, we train a version of the \drmd agent using the transitions and reward structure of ICMDP~\cite{lin2020deep}. Concretely, episodes continue until either a) malware is misclassified or b) the agent runs out of training samples. Otherwise, states and actions are the same as in the MD-MDP. As a result, the next states are based on the next sample from the dataset. Rewards are structured as in $R_{acc}$, however, rewards from goodware samples are scaled down by $\hat\sigma$.
            For fairness, we use the same PPO architecture and hyperparameters to train both our agent and the ICMDP agent.
            The $\Delta IC$ columns in Table~\ref{tab:post-hoc-results} show the gains or losses of the MD-MDP \drmd agent over the ICMDP \drmd agent across all combinations of dataset-feature-space pairs, AL rates, and rejection rates. Overall, the MD-MDP outperforms ICMDP in $97$ out of $100$ settings, with 45 settings being statistically significant, and achieves an average $\Delta IC$ of $1.94$. These results demonstrate that beyond theoretical modeling discussions, the MD-MDP is empirically better suited for integration into wider malware detection pipelines.

    \paragraph{One-Step PPO vs DCBs}
        To isolate the benefit of our architecture and algorithm choice, we compare two versions of \drmd. The first is the standard version of \drmd and the second uses one of two DCBs, either NeuralTS or NeuralUCB, depending on which performs better for a given setting. The DCBs use the same architecture and algorithms as proposed in their original works. The $\Delta$DCB columns in Table~\ref{tab:post-hoc-results} shows the gains of one-step PPO over the DCBs across all combinations of dataset-feature-space pairs, AL rates, and rejection rates. Overall, the one-step PPO shows statistically significant improvements over the DCBs in all $100$ settings. These results demonstrate that the architecture and one-step PPO algorithm chosen for \drmd is better suited, than standard DCBs, to the dynamics of the Android malware detection considered in this paper.
        
    \begin{table*}[t]
    \centering
    \caption{\textbf{Ablation Study. }\drmd performance and rejection/selection rates starting with the simplest base approach and adding components of \drmd back sequentially. AUTs reported show the mean $\pm$ std performance of \drmd across five runs after adding the component, specified at the start of the row, to the version of \drmd from the previous row. '---' denotes rejection is not enabled and italics denote that the samples are also used for AL.}
    \label{tab:ablation}
    \resizebox{1\textwidth}{!}{
    \begin{tabular}{l|lr|lr|lr|lr}
        \toprule
        \multirow{2}{*}{Added} & \multicolumn{2}{c|}{Hypercube-Drebin} & \multicolumn{2}{c|}{Hypercube-Ramda} & \multicolumn{2}{c|}{Transcendent-Drebin} & \multicolumn{2}{c}{Transcendent-Ramda} \\
         & \multicolumn{1}{l}{AUT $\pm$ std} & \multicolumn{1}{r|}{Rejected} & \multicolumn{1}{l}{AUT $\pm$ std} & \multicolumn{1}{r|}{Rejected} & \multicolumn{1}{l}{AUT $\pm$ std} & \multicolumn{1}{r|}{Rejected} & \multicolumn{1}{l}{AUT $\pm$ std} & \multicolumn{1}{r}{Rejected} \\
        \midrule
        Basic DRMD & 59.22 $\pm$ 1.58 & --  &  22.34 $\pm$ 1.52 & --  &  46.53 $\pm$ 1.48 & --  &  37.75 $\pm$ 2.21 & -- \\ 
        Temporal Reward Scaling & 56.15 $\pm$ 0.64 & --  &  25.74 $\pm$ 1.55 & --  &  46.18 $\pm$ 2.07 & --  &  37.39 $\pm$ 2.44 & -- \\ 
        Malware Reward Scaling & 64.07 $\pm$ 1.77 & --  &  43.95 $\pm$ 0.88 & --  &  56.80 $\pm$ 0.65 & --  &  49.80 $\pm$ 1.90 & -- \\ 
        4 Hidden Network Layers & 63.56 $\pm$ 1.69 & --  &  43.44 $\pm$ 1.43 & --  &  57.69 $\pm$ 0.83 & --  &  49.90 $\pm$ 0.56 & -- \\ 
        512 Neuron Hidden Layers & 63.33 $\pm$ 1.14 & --  &  45.03 $\pm$ 0.68 & --  &  58.07 $\pm$ 0.53 & --  &  50.46 $\pm$ 0.87 & -- \\ 
        \midrule
        Integrated Reject Action & 61.42 $\pm$ 0.90 & 0.00  &  44.90 $\pm$ 0.79 & 0.00  &  57.88 $\pm$ 2.22 & 0.00  &  50.28 $\pm$ 1.22 & 0.00 \\ 
        Reward Reject Outcome & 61.68 $\pm$ 1.71 &  52.02  &  50.54 $\pm$ 2.82 &  533.80  &  60.70 $\pm$ 2.34 &  217.31  &  56.90 $\pm$ 0.74 &  599.47 \\ 
        Reject Cost ($R_{cost}$=-0.1) & 63.17 $\pm$ 1.27 &  58.17  &  50.99 $\pm$ 4.94 &  529.27  &  58.51 $\pm$ 4.34 &  203.50  &  55.84 $\pm$ 1.09 &  595.97 \\ 
        Integrated Rejection with AL & 64.24 $\pm$ 1.78 &  \textit{9.70}  &  52.49 $\pm$ 1.07 &  \textit{391.02}  &  59.47 $\pm$ 0.79 &  \textit{17.11}  &  62.90 $\pm$ 1.07 &  \textit{170.39} \\ 
        Sliding Retraining Window & 67.51 $\pm$ 5.41 &  \textit{21.99}  &  65.39 $\pm$ 1.82 &  \textit{956.98}  &  60.39 $\pm$ 8.24 &  \textit{20.98}  &  68.32 $\pm$ 6.43 &  \textit{869.61} \\ 
        Augmented Active Learning & 77.48 $\pm$ 1.95 &  \textit{400}$|$13.83  &  59.61 $\pm$ 0.64 &  \textit{400}$|$684.16  &  74.57 $\pm$ 0.56 &  \textit{400}$|$64.27  &  66.82 $\pm$ 1.49 &  \textit{400}$|$401.47 \\ 
        \bottomrule
    \end{tabular}
    }
\end{table*}

    \subsection{Classification-Rejection Policy}
        Extending from the Classification-only policy, the \drmd Classification-Rejection policy ($\pi_\Sigma$) integrates the \emph{rejection action}, $a_{rej}$. Importantly, this action enables the agent to abstain from making a decision on a sample if there is insufficient information, high uncertainty, or concept drift. Note how, previously, rejections occurred only at the end of each month; however, the rejection action provides \emph{real-time} abstention as the samples are seen, thus opening up the potential for real-time detection systems in future work. To maintain the capability for such decision making while facilitating AL, the samples resulting from the rejection action are used for retraining. We consider three different rejection costs ($R_{cost}$) to show how they effect the performance of DRL agents that can reject malware samples, which has not been considered in prior work. We present the results in Table~\ref{tab:integrated-rejection}, showing $\Delta Base$ as the comparison of \drmd to the best baseline. To ensure consistency in the number of rejections, the seed-wise average rejections (rounded up) of \drmd is used as the budget for baseline rejections.
        
        \paragraph{Integrated Rejection (IR)}
        The IR setting introduces the rejection action without any additional AL on rejected samples. The results can be seen in Table~\ref{tab:integrated-rejection}, under `\textit{IR}', and show statistically significant improvements over all baselines in $9$ of the $12$ settings with an average $\Delta Base$ of $16.15$. This shows that the agent can effectively learn a policy that rejects samples likely to be misclassified. When comparing the two feature spaces (Ramda and Drebin), we see a disparity in the average number of rejections. Specifically, policies trained on Ramda reject more samples than those trained on Drebin. We hypothesize that the reduction of information encapsulated by Ramda results in greater prediction uncertainty compared to the Drebin feature space. Increasing $R_{cost}$ intuitively reduces the number of rejected samples with only a minimal effect on $\Delta Base$. This implies that higher $R_{cost}$ values lead \drmd to be more selective in rejection. 

        \paragraph{IR with Active Learning (IRAL)} 
        To go beyond the mitigation of concept drift and allow self-adaptation, we include the samples rejected in each period for retraining \drmd by updating the fine-tuning sample sliding window. The IRAL section in Table~\ref{tab:integrated-rejection} shows the results of the combined rejection and AL policy across each combination of the dataset-feature-space-pair per rejection cost. Compared with the best baseline for each setting, \drmd shows statistically significant improvements in $7$ of the $12$ settings with an average $\Delta Base$ of $11.60$. In Hypercube-Drebin and Transcendent-Drebin, retraining on rejected samples leads to higher confidence in classification actions. As $R_{cost}$ increases, this leads to a reduction in the number of rejected and, thus, retraining samples in subsequent periods causing the observed drop in performance.
        
        \paragraph{IR with Augmented Active Learning (IRAAL)}
        To help mitigate the impact of conservative rejection policies and ensure stable retraining budgets for AL, the IRAAL section in Table~\ref{tab:integrated-rejection} presents integrated rejection with augmented AL. Augmented AL extends the rejection-based sampling used in IRAL with uncertainty sampling so that, given a predetermined budget for AL, the model samples up to the predetermined labeling budget for retraining if the agent was conservative in rejections, and downsamples to the budget if the agent is rejecting more samples. We compare using the four different AL budgets as in Section~\ref{sec:class_only_pol}. The results show statistically significant performance gains over the baseline approaches in 33 of 48 settings, with an average $\Delta Base$ of $10.75$. Highlighting that the augmented AL budget improves stability and performance compared to the IRAL results.

    \paragraph{Ablation Study}
        In Table~\ref{tab:ablation} we present a cumulative component breakdown of the AUT performance of \drmd across the four dataset–feature‐space pairs.
        Starting from the \textit{basic \drmd} (single 128‐neuron layer, ±1 reward, no rejection), performance is uniformly low. Introducing \textit{temporal and malware reward scaling} immediately increases AUT across all settings, highlighting the importance of incorporating both the spatial (malware distribution) and temporal aspects of the domain into the reward function. Increasing the capacity of the model, both by the number of \textit{hidden layers} and by the number of their \textit{neurons}, provides marginal improvements.
        Adding the \textit{integrated reject action} without reward $R_{rej}$ leads to the model not rejecting any samples and to negligible performance differences. Crucially, \textit{rewarding rejected outcomes} provides a large performance increase as it allows the agent to learn to balance abstention and misclassification risk, with the \textit{rejection cost} further refining this trade‐off. 
        Subsequently, \textit{integrated rejection with AL} enables the agent to adapt to drift over time and coupled with the \emph{sliding retraining window} provides performance increases in all but the Transcendent-Drebin setting where \drmd becomes too conservative in rejection. By rectifying this, \emph{augmented active learning} yields a increase in AUT performance across most settings.

\section{Conclusion} \label{sec:conclusion}
We introduce a new one-step MDP formulation for malware detection, MD-MDP, which enables the unification of DRL, cost-aware rejection, and AL to counteract concept drift in temporally split evaluations. We use MD-MDP to train the \drmd agent. 
Our results demonstrate that across two Android malware datasets and two feature spaces, \drmd outperforms the baselines of Drebin, DeepDrebin, Ramda, and SL-DRMD (SL version of \drmd) by an average of $8.66$ AUT using the classification-only policy. 
Furthermore, we show that \drmd learns effective strategies that enable real-time detection, rejection, and AL leading to an average $10.90$ AUT increase over the best baselines for all settings using the classification-rejection policy.

Future work could extend the MD-MDP and \drmd to include additional aspects of the malware detection pipeline, such as feature retrieval, as done in the image domain by Janisch \etal~\citeyear{janisch2019classification}. Additionally, extending the rejection reward design to integrate sources of information, such as conformal evaluation~\cite{Transcend,Transcendent}, explanation~\cite{CADE}, or contrastive learning~\cite{chen2023continuous}, provides a potentially valuable next step. Together these directions for future work represent the path forward to a DRL policy that unifies the complete Android malware detection problem.

\section*{Acknowledgments}
This research was partially funded and supported by: 
the UK National Cyber Security Centre (NCSC);
UK EPSRC Grant no. EP/X015971/2;
the Defence Science and Technology Laboratory (DSTL), an executive agency of the UK Ministry of Defence, supporting the Autonomous Resilient Cyber Defence (ARCD) project within the DSTL Cyber Defence Enhancement programme.

\bibliographystyle{plainnat}
\bibliography{ref}

\appendix

\section{Computing Infrastructure} \label{app:infra}
All experiments were conducted on a server using x86\_64 architecture, AMD EPYC processors (16 CPU cores total), a NVIDIA H100 NVL GPU (CUDA 12.6) with 96 GB of memory, and 128 GB of RAM. The evaluation used Python 3.9.21 with core libraries including PyTorch 2.4.1+cu121, TensorFlow 2.18.0, scikit-learn 1.6.1, and NumPy 2.0.2. Further details regarding all library requirements can be found in the repository provided.

\section{Seeding} \label{app:seeding}
In order to improve the reproducibility of the experiments presented in the paper, the random seeds presented in Appendix~\ref{app:hyper} are used to seed $random$, $numpy.random$, and $torch.manual\_seed$.

\section{Hyperparameters} \label{app:hyper}
This section presents the complete hyperparameters and architecture for all approaches used in this paper.

    \subsection{Drebin}
        The hyperparameters used for all Drebin experiments was drawn directly from the updated time-aware evaluation work by Kan \etal~\citeyear{kan2024tesseract}. The \textbf{hyperparameters} used are as follows, \textit{Max Iterations}: 50000, \textit{C}: 1, \textit{Seeds}: [0, 1, 26, 42, 0x10c0ffee], \textit{Classifier}: LinearSVC (sklearn.svm).

    \subsection{DeepDrebin}
        The hyperparameters used for all DeepDrebin experiments were drawn directly from the updated time-aware evaluation work by Kan \etal~\citeyear{kan2024tesseract}. The \textbf{hyperparameters} used are as follows:
        
        \textbf{Training}, \textit{Epochs}: 10, \textit{Batch Size}: 64, \textit{Training/Validation Split}: 0.66/0.34, \textit{Seed}: [0, 1, 26, 42, 0x10c0ffee], \textit{Learning Rate}: 0.05, \textit{Loss}: Cross Entropy, \textit{Optimizer}: SGD. 
        
        \textbf{Architecture}, \textit{Input}: n\_features, \textit{Layer Size}: 200, \textit{Hidden Layers}: 2, \textit{Activation Function}: ReLU, \textit{Dropout}: 0.5, \textit{Output}: 2.

    \subsection{Ramda}
    The hyperparameters used for Ramda follow the original work by Heng \etal~\citeyear{ramda}, extended to match our setting and hyperparameters used with the rest of the baselines. The \textbf{hyperparameters} used are as follows:
    
    \textbf{Training}, \textit{Epochs}: 10, \textit{Batch Size}: 512, \textit{Training/Validation Split}: 0.80/0.20, \textit{Seeds}: [0, 1, 26, 42, 0x10c0ffee], \textit{Learning Rate}: 1e-3, \textit{Optimizer Epsilon}: 1e-8, \textit{Optimizer}: Adam, \textit{Reconstruction Weight}: 10.0, \textit{KL Divergence Weight}: 1.0, \textit{Feature Disentanglement Weight}: 10.0.

    \textbf{VAE}, \textit{Input}: n\_features, \textit{Hidden Layer Size}: 600, \textit{Dropout}: 0.1, \textit{Encoder Hidden Layers}: 2, \textit{Bottleneck Layer Size}: 80,
    \textit{Decoder Hidden Layers}: 2, \textit{Activation Function}: ELU (first layer), Tanh (second layer) for encoder and reversed for decoder, \textit{Input}: n\_features, \textit{V Threshold}: 30.

    \textbf{MLP}, \textit{Input}: 2 $\times$ Bottleneck Layer Size, \textit{Layer Size}: 40, \textit{Hidden Layers}: 3, \textit{Activation Function}: Tanh (first layer) and ELU (second \& third layers), \textit{Dropout}: 0.1, \textit{Output}: 2. 

    \subsection{DRMD}
        The final hyperparameters used for \drmd are the same used in development, with the hyperparameters following from Schulman \etal~\citeyear{schulman2017proximal} and Huang \etal~\citeyear{huang2022cleanrl}. The \textbf{hyperparameters} used are as follows:
        
        \textbf{Training}, \textit{Epochs (Training Data)}: 5, \textit{Epochs (Minibatches)}: 1, \textit{Sliding Window Size}: 5000, \textit{Minibatch Size}: 100, \textit{Clip Coefficient}: 0.2, \textit{Value Coefficient}: 0.5, \textit{Entropy Coefficient}: 0.01, \textit{Max Grad Norm}: 0.5, \textit{Seeds}: [0, 1, 26, 42, 0x10c0ffee], \textit{Learning Rate}: 2.5e-4, \textit{Optimizer Epsilon}: 1e-5, \textit{Optimizer}: Adam. 
        
        \textbf{Actor}, \textit{Input}: n\_features, \textit{Layer Size}: 512, \textit{Hidden Layers}: 4, \textit{Activation Function}: LeakyReLU, \textit{Dropout}: 0.5, \textit{Output}: 3. 
        
        \textbf{Critic}, \textit{Input}: n\_features, \textit{Layer Size}: 512, \textit{Hidden Layers}: 4, \textit{Activation Function}: LeakyReLU, \textit{Dropout}: 0.5, \textit{Output}: 1.

    \subsection{SL-DRMD}
        The hyperparameters of SL-DRMD are a combination of the DeepDrebin and \drmd hyperparameters. The \textbf{hyperparameters} used are as follows:
        
        \textbf{Training}, \textit{Epochs}: 5, \textit{Batch Size}: 256, \textit{Training/Validation Split}: 0.66/0.34, \textit{Seeds}: [0, 1, 26, 42, 0x10c0ffee], \textit{Learning Rate}: 0.05, \textit{Loss}: Cross Entropy, \textit{Optimizer}: SGD. 
        
        \textbf{Architecture}, \textit{Input}: n\_features, \textit{Layer Size}: 512, \textit{Hidden Layers}: 4, \textit{Activation Function}: ReLU, \textit{Dropout}: 0.5, \textit{Output}: 2.
    
    \subsection{ICMDP}
        We use the same hyperparameters for the ICMDP version architecture as our proposed \drmd to isolate the modeling difference between the two versions, showing the impact of each version of the reward and state transition functions. We report the Gamma and GAE Lambda hyperparameters as they are relevant for the ICMDP version because of its multi-step episodes. The \textbf{hyperparameters} used are as follows:
        
        \textbf{Training}, \textit{Epochs (Training Data)}: 5, \textit{Epochs (Minibatches)}: 1, \textit{Sliding Window Size}: 5000, \textit{Minibatch Size}: 100, \textit{Clip Coefficient}: 0.2, \textit{Value Coefficient}: 0.5, \textit{Entropy Coefficient}: 0.01, \textit{Max Grad Norm}: 0.5, \textit{Gamma}: 0.99, \textit{GAE Lambda}:  0.95, \textit{Seeds}: [0, 1, 26, 42, 0x10c0ffee], \textit{Learning Rate}: 2.5e-4, \textit{Optimizer Epsilon}: 1e-5, \textit{Optimizer}: Adam. 
        
        \textbf{Actor}, \textit{Input}: n\_features, \textit{Layer Size}: 512, \textit{Hidden Layers}: 4, \textit{Activation Function}: LeakyReLU, \textit{Dropout}: 0.5, \textit{Output}: 2. 
        
        \textbf{Critic}, \textit{Input}: n\_features, \textit{Layer Size}: 512, \textit{Hidden Layers}: 4, \textit{Activation Function}: LeakyReLU, \textit{Dropout}: 0.5, \textit{Output}: 1.

    \subsection{NeuralUCB \& NeuralTS}
    The hyperparameters used forNeuralTS~\cite{zhang2020neural} and NeuralUCB~\cite{zhou2020neural} follow their original works, extended to match our setting. The \textbf{hyperparameters} used are as follows:

    \textbf{Training}, \textit{Gradient Steps per Update}: 100, \textit{Minibatch Size}: $|\mathcal{D}|$ (full batch over stored contexts), \textit{Loss}: MSE, \textit{Seeds}: [0, 1, 26, 42, 0x10c0ffee], \textit{Learning Rate}: 1e-2, \textit{Weight Decay}: $\nicefrac{\lambda}{|\mathcal{D}|}$, \textit{Optimizer}: SGD. 

    \textbf{Arm Architecture}, \textit{Input}: 2 $\times$ n\_features (block-encoded context), \textit{Layer Size}: 100, \textit{Hidden Layers}: 1, \textit{Activation Function}: ReLU, \textit{Output}: 1.

    \textbf{Bandit Parameters}, \textit{Arms}: 2, \textit{Style}: TS (NeuralTS) or UCB (NeuralUCB), ($\lambda$): 1, ($\nu$): 1, \textit{Gradient Estimation}: BackPACK.

\end{document}